\documentclass[journal, 10pt, twoside]{IEEEtran}
\usepackage{cite}
\usepackage{graphicx}
\usepackage{multirow}
\usepackage{url}
\usepackage{amssymb}
\usepackage[dvipsnames, svgnames, x11names]{xcolor}
\usepackage{colortbl}
\usepackage{footnote}
\usepackage{scalerel}
\usepackage{tikz}
\usetikzlibrary{svg.path}
\definecolor{orcidlogocol}{HTML}{A6CE39}
\tikzset{
  orcidlogo/.pic={
    \fill[orcidlogocol] svg{M256,128c0,70.7-57.3,128-128,128C57.3,256,0,198.7,0,128C0,57.3,57.3,0,128,0C198.7,0,256,57.3,256,128z};
    \fill[white] svg{M86.3,186.2H70.9V79.1h15.4v48.4V186.2z}
                 svg{M108.9,79.1h41.6c39.6,0,57,28.3,57,53.6c0,27.5-21.5,53.6-56.8,53.6h-41.8V79.1z M124.3,172.4h24.5c34.9,0,42.9-26.5,42.9-39.7c0-21.5-13.7-39.7-43.7-39.7h-23.7V172.4z}
                 svg{M88.7,56.8c0,5.5-4.5,10.1-10.1,10.1c-5.6,0-10.1-4.6-10.1-10.1c0-5.6,4.5-10.1,10.1-10.1C84.2,46.7,88.7,51.3,88.7,56.8z};
  }
}
\newcommand{\settablefont}{\fontsize{8.0}{12}\selectfont}
\newcommand\orcidicon[1]{\href{https://orcid.org/#1}{\mbox{\scalerel*{
\begin{tikzpicture}[yscale=-1,transform shape]
\pic{orcidlogo};
\end{tikzpicture}
}{|}}}}
\usepackage{hyperref}
\hypersetup{
    colorlinks=true,
    linkcolor=blue,
    filecolor=black,      
    urlcolor=black,
    citecolor=purple
}
\usepackage{tikz-network}
\usepackage[switch]{lineno}

\newcommand\normx[1]{\left\Vert#1\right\Vert}

\begin{document}
\title{Three-Filters-to-Normal+: Revisiting Discontinuity Discrimination in Depth-to-Normal Translation}
\author{Jingwei Yang$^{\orcidicon{0000-0003-2835-4962}\,}$,~\IEEEmembership{Graduate Student Member,~IEEE}, Bohuan Xue$^{\orcidicon{0000-0003-0332-3967}\,}$,~\IEEEmembership{Graduate Student Member,~IEEE}, 
\\
Yi Feng$^{\orcidicon{https://orcid.org/0009-0005-4885-0850}\,}$,~\IEEEmembership{Graduate Student Member,~IEEE}, Deming Wang$^{\orcidicon{0000-0003-3486-4176}\,}$,~\IEEEmembership{Member,~IEEE}, \\
\ \ \
Rui Fan$^{\orcidicon{0000-0003-2593-6596}\,}$,~\IEEEmembership{Senior Member,~IEEE}, Qijun Chen$^{\orcidicon{0000-0001-5644-1188}\,}$,~\IEEEmembership{Senior Member,~IEEE}
\thanks{
This research was supported by the National Key R\&D Program of China under Grant 2020AAA0108100, the National Natural Science Foundation of China under Grant 62233013, the Science and Technology Commission of Shanghai Municipal under Grant 22511104500, and the Fundamental Research Funds for the Central Universities.
(\emph{Jingwei Yang and Bohuan Xue share first authorship}) (\emph{Corresponding author: Rui Fan}).
}
\thanks{Jingwei Yang, Yi Feng, Deming Wang, Rui Fan, and Qijun Chen are with the College of Electronics \& Information Engineering, Shanghai Research Institute for Intelligent Autonomous Systems, the State Key Laboratory of Intelligent Autonomous Systems, and Frontiers Science Center for Intelligent Autonomous Systems, Tongji University, Shanghai 201804, China. (e-mails: \{jw\_yang, fengyi0109, wangdeming, rfan, qjchen\}@tongji.edu.cn)
}
\thanks{Bohuan Xue is with the Department of Computer Science \& Engineering, the Hong Kong University of Science and Technology, Hong Kong SAR, China. (e-mail: bxueaa@connect.ust.hk)
}
}
	
\maketitle

\begin{abstract}
This article introduces three-filters-to-normal+ (3F2N+), an extension of our previous work three-filters-to-normal (3F2N), with a specific focus on incorporating discontinuity discrimination capability into surface normal estimators (SNEs). 3F2N+ achieves this capability by utilizing a novel discontinuity discrimination module (DDM), which combines depth curvature minimization and correlation coefficient maximization through conditional random fields (CRFs). To evaluate the robustness of SNEs on noisy data, we create a large-scale synthetic surface normal (SSN) dataset containing 20 scenarios (ten indoor scenarios and ten outdoor scenarios with and without random Gaussian noise added to depth images). Extensive experiments demonstrate that 3F2N+ achieves greater performance than all other geometry-based surface normal estimators, with average angular errors of 7.85$^\circ$, 8.95$^\circ$, 9.25$^\circ$, and 11.98$^\circ$ on the clean-indoor, clean-outdoor, noisy-indoor, and noisy-outdoor datasets, respectively. We conduct three additional experiments to demonstrate the effectiveness of incorporating our proposed 3F2N+ into downstream robot perception tasks, including freespace detection, 6D object pose estimation, and point cloud completion. Our source code and datasets are publicly available at \url{https://mias.group/3F2Nplus}.
\\

\textit{Note to Practitioners}---The primary motivation behind this work arises from the need to develop a high-performing surface normal estimator for practical robotics and computer vision applications. While geometry-based surface normal estimators have been widely used in these domains, the existing solutions focus merely on discontinuity discrimination. To tackle this problem, this article introduces a plug-and-play module that leverages both depth curvature and correlation coefficient to quantify discontinuity levels, thereby optimizing surface normal estimation, particularly near or on discontinuous regions. Moreover, this article also introduces a large-scale public dataset with random noise added to depth images, providing a more realistic and robust platform for algorithm evaluation within this research community. Extensive experimental results demonstrate that our method outperforms other state-of-the-art algorithms.
\end{abstract}


\section{Introduction}
\label{sec:introduction}

\IEEEPARstart{A}S an informative visual feature representing planar characteristics, surface normal has been prevalently utilized in numerous computer vision and robotics applications, such as collision-free space detection \cite{fan2020sne, li2023roadformer, wang2021sne, wang2021dynamic}, point cloud completion \cite{huang2020pf}, and 6D object pose estimation \cite{liu2023bdr6d}. Existing surface normal estimators (SNEs) have predominantly employed optimization techniques \cite{klasing2009comparison,holzer2012adaptive,badino2011fast, zeisl2014discriminatively}, \textit{e.g.}, singular value decomposition (SVD) \cite{kalman1996singularly} and principal component analysis (PCA) \cite{abdi2010principal} to process unstructured range sensor data. Such algorithms are, nevertheless, computationally intensive in nature \cite{fan2021three}. Therefore, there has been a shift in focus towards structured range sensor data, specifically depth images. This category of algorithms is typically referred to as ``\textit{depth-to-normal translator}''. Our previously published works, \textit{e.g.}, three-filters-to-normal (3F2N) \cite{fan2021three}, depth-to-normal translator (D2NT) \cite{d2nt2023}, and spatial discontinuity-aware SNE (SDA-SNE) \cite{ming2022sda}, leverage basic image filters to perform depth-to-normal translation. Such approaches stand as the state-of-the-art (SoTA) in the domain of surface normal estimation.

\begin{figure*}[t!]
    \centering
		\includegraphics[width=0.9999\linewidth]{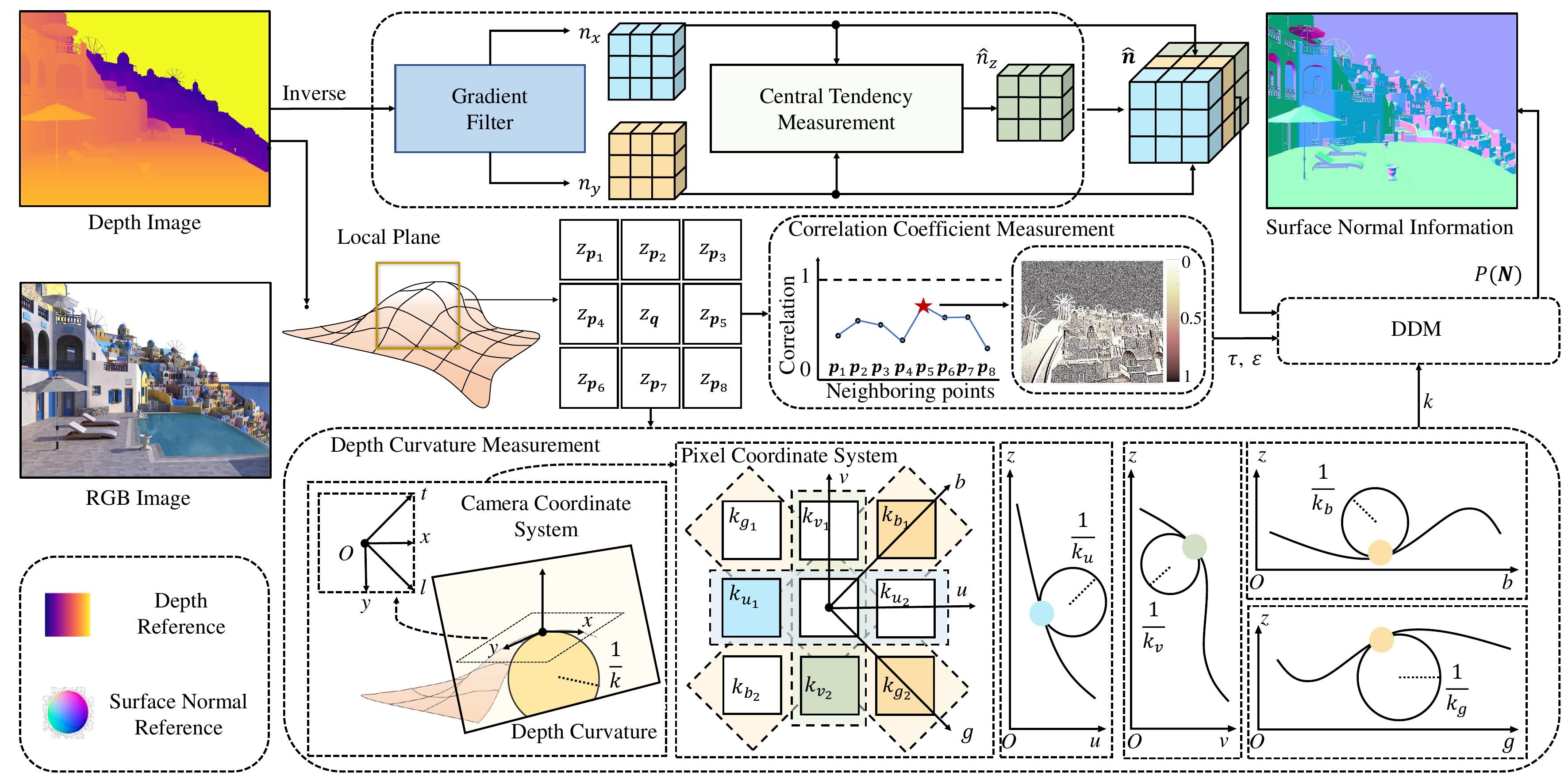}
		\caption{The workflow of our proposed 3F2N+ SNE. 
            }
		\label{frame}
\end{figure*}

The performance of an SNE is profoundly impacted by the quality of the raw sensor data (point clouds or depth maps) it processes \cite{d2nt2023}. Such sensor data often exhibit substantial variations, especially in regions with discontinuities \cite{d2nt2023}. Therefore, the primary objective of this study is to introduce innovative approaches to optimize surface normals in these discontinuous regions. Additionally, it is important to note that for real-world datasets, such as NYU2 \cite{silberman2012indoor} and KITTI \cite{geiger2012are}, surface normal ground truth is often unavailable \cite{d2nt2023}. Their so-called ``ground truth'' is derived by estimating surface normals directly from noisy depth images or 3D point clouds using a traditional SNE. While we have provided a public dataset alongside 3F2N in \cite{fan2021three}, this dataset is devoid of any noise and cannot accurately reflect the performance of SNEs in realistic scenarios.

To address the existing challenges mentioned above, our primary focus in this article is to enhance the discontinuity discrimination capability of our previous work 3F2N. The main contributions of this work can be summarized as follows:
\begin{itemize}
    \item An exploration into the feasibility of utilizing depth curvature and correlation coefficient for discontinuity extent quantification.
    \item A novel discontinuity discrimination module (DDM), which incorporates the measured discontinuity extent into the feature function of conditional random field (CRF), effectively distinguishing discontinuities and thereby optimizing surface normal estimation. The 3F2N integrated with DDM is denoted as 3F2N+ in this article.
    \item A large-scale synthetic surface normal (SSN) dataset consisting of 20K depth images (both clean and noisy) and their corresponding surface normal ground truth. This dataset is designed to comprehensively evaluate SNE performance in both indoor and outdoor scenarios, with an even split of ten scenarios each.
    \item A series of additional experiments underscore the benefits of incorporating our proposed 3F2N+ SNE into three downstream computer vision and robotics tasks: (1) data-fusion freespace detection, (2) 6D object pose estimation, and (3) point cloud completion.
\end{itemize}

The remainder of this article is structured as follows: Sect. \ref{sec:rw} provides a comprehensive review of the SoTA SNEs. Sect. \ref{sec:methodology} introduces our proposed 3F2N+ SNE. Extensive experiments demonstrating the robustness of our approach are detailed in Sect. \ref{sec.exp}. Sect. \ref{sec.app} delves into a range of computer vision and robotics applications assisted with 3F2N+ SNE. Finally, we summarize our work in Sect. \ref{sec.conclusion}.

\section{Literature Review}
\label{sec:rw}
In this section, we provide a comprehensive overview of the SoTA surface normal estimators. A 3D point is denoted as $\boldsymbol{q}=(x; y; z)$, and its corresponding surface normal is denoted as $\boldsymbol{n}=(n_x; n_y; n_z)$. A set storing the neighboring points of $\boldsymbol{q}$ is denoted as $\boldsymbol{P}=(\boldsymbol{p}_{1}, ..., \boldsymbol{p}_{k})$. An augmented set of $\boldsymbol{P}$ is denoted as $\boldsymbol{P}^+=(\boldsymbol{q}, \boldsymbol{P})$.
\subsection{Point Cloud-to-Normal SNEs}
    In this category of SNEs\footnote{Point cloud-to-normal approaches can also be employed to estimate surface normals from depth images when the camera intrinsic matrix is known.}, surface normals are computed from 3D point clouds by determining the normal vectors of locally interpolated planar surfaces. One common approach formulates this process as the following optimization problem:
    \begin{equation}
		\begin{split}
                \underset{\boldsymbol{n}}{\arg\min} \ J(\boldsymbol{q}, \boldsymbol{P}, \boldsymbol{n}),
		\end{split}\label{con:optimization}
    \end{equation}
    where $J(\cdot)$ denotes the cost function, which takes various criteria into account depending on the specific method being used. For instance, it may quantify the plane fitting error in \textit{PlaneSVD} \cite{p6} and \textit{PlanePCA} \cite{klasing2009realtime}, the angle difference in \textit{VectorSVD} \cite{klasing2009comparison}, or the plane fitting error in the spherical coordinate system in \textit{FALS} \cite{badino2011fast}. Another solution to this problem is to average the surface normals from neighboring triangles, which can be formulated as follows:
    \begin{equation}
	\begin{split}
            \boldsymbol{n}=\frac{1}{k} \sum\limits_{i=1}^{k} \omega_i \frac{ [\boldsymbol{p}_{i}-\boldsymbol{q}]_{\times} (\boldsymbol{p}_{i+1}-\boldsymbol{q})}{
            \normx{[\boldsymbol{p}_{i}-\boldsymbol{q}]_{\times} ( \boldsymbol{p}_{i+1}-\boldsymbol{q})}_2
            },
	\end{split}\label{con:weighted}
    \end{equation}
    where $\omega_i$ denotes the weight associated with the given triangle. \textit{AreaWeighted} \cite{jin2005comparison} determines $\omega_i$ based on the magnitude of each triangle, while \textit{AngleWeighted} \cite{jin2005comparison} determines $\omega_i$ based on the angle between vectors $\boldsymbol{p}_{i}-\boldsymbol{q}$ and $\boldsymbol{p}_{i+1}-\boldsymbol{q}$. \textit{SRI} \cite{badino2011fast} computes surface normals using the following expression:
    \begin{equation}
		\begin{split} 
        \boldsymbol{n}=(\hat{\boldsymbol{z}};\hat{\boldsymbol{x}};\hat{\boldsymbol{y}})\boldsymbol{R}(1; \frac{\partial r}{r\cos\phi\partial\theta}; \frac{\partial r}{r\partial\phi}),
		\end{split}\label{con:sri}
    \end{equation}
     where $\hat{\boldsymbol{x}}$, $\hat{\boldsymbol{y}}$, and $\hat{\boldsymbol{z}}$ denote the unit vectors along the respective coordinate directions, respectively, and $r$, $\theta$, and $\phi$ denote the range, azimuth, and elevation components in the spherical coordinate system, respectively, and 
     $\boldsymbol{R}$ is a rotation matrix.
    
\subsection{Depth-to-Normal SNEs}
   \textit{LINE-MOD} \cite{hinterstoisser2011gradient} is a pioneering depth-to-normal SNE. It starts by computing the optimal depth gradient and then forms a 3D plane using three triangle vertices. Finally, the surface normal can be obtained as the tangential vector of the 3D plane. In our previous works 3F2N \cite{fan2021three}, SNE-RoadSeg \cite{fan2020sne}, and SNE-RoadSeg+ \cite{wang2021sne}, we utilize basic image gradient filters to compute the components $n_x$ and $n_y$ of surface normals: 
    \begin{equation}
		\begin{split}
		n_x=f_x\frac{\partial 1/z}{\partial u}=f_x g_u, \ \ n_y=f_y\frac{\partial 1/z}{\partial v}=f_y g_v, 
		\end{split}\label{con:3f2n_nxny}
    \end{equation}
    where $f_x$ and $f_y$ denote the camera focal lengths (in pixels) in the $x$ and $y$ directions, respectively. SNE-RoadSeg computes $n_z$ using the following expression \cite{fan2019pothole}: 
    \begin{equation}
	n_z=\cos(\arctan\frac{\sum_{i=1}^{k}\Bar{n}_{x,i}\cos\phi+\sum_{i=1}^{k}\Bar{n}_{y,i}\sin\phi}{\sum_{i=1}^{k}\Bar{n}_{z,i}}),
	\label{con:roadsegphi}
    \end{equation}
    where $\bar{\boldsymbol{n}}_i=\frac{\boldsymbol{n}_i}{||\boldsymbol{n}_i||_2}=(\bar{n}_{x,i};\bar{n}_{y,i};\bar{n}_{z,i})$ and $\phi=\arctan(\frac{f_yg_v}{f_xg_u})$. 3F2N computes $n_z$ using the following expression: 
    \begin{equation}
	\begin{split}
	    n_z=-\Phi\left\lbrace\frac{\Delta{x_{i}}n_x+\Delta{y_{i}}n_y}{\Delta{z_{i}}}\right\rbrace, 
	\end{split}\label{con:3f2n_nz}
    \end{equation}
    where $\Phi\{\cdot\}$ denotes a central tendency measurement approach and $(\Delta{x_{i}};\Delta{y_{i}};\Delta{z_{i}})=\boldsymbol{p}_{i}-\boldsymbol{q}$. Our recent works D2NT \cite{d2nt2023} and SDA-SNE \cite{ming2022sda} compute surface normals in an end-to-end manner as follows: 
       \begin{equation}
		\begin{split}
        \boldsymbol{n}_i=(f_x\frac{\partial z}{\partial u};f_y\frac{\partial z}{\partial v};-z-\frac{(u-u_0)\partial z}{\partial u}-\frac{(v-v_0)\partial z}{\partial v}),
		\end{split}\label{con:d2nt}
    \end{equation}
    where $\boldsymbol{p}_0=(u_0;v_0)$ is the image principal point.

\section{Methodology\label{sec:methodology}}
The framework is schematically illustrated in Fig. \ref{frame}. A $3\times3$ local region contains the depth $z_{\boldsymbol{q}}$ of the target point $\boldsymbol{q}$, along with depths $z_{\boldsymbol{p}_i}$ of the eight neighboring points $\boldsymbol{p}_i$ $(i=[1,8]\bigcap\mathbb{Z}^+)$. We introduce two methods to enhance the accuracy of surface normal estimation near/on discontinuities. One is based on the depth curvature minimization (see Sect. \ref{sec:curvation}), while the other is based on the correlation coefficient maximization (see Sect. \ref{sec:corelation}). Finally, we present an optimization strategy (see Sect. \ref{sec:ddm}) that integrates these two modules and can be embedded into all SNEs.

\subsection{Depth Curvature-Based Discontinuity Discrimination}
\label{sec:curvation}
    In this subsection, we aim to quantify local surface smoothness using the depth curvature $k$ of neighboring points. A lower depth curvature corresponds to a smoother surface \cite{federer1959curvature}. To comprehensively discriminate discontinuities, we calculate four types of curvatures: mean, maximum, normal, and Gaussian. Mean and maximum depth curvatures are typically obtained by computing the first- and second-order partial derivatives of depth. Therefore, our initial step involves computing the first-order partial derivatives of depth in both the $x$ and $y$ directions within the camera coordinate system as follows:
    \begin{equation}
        \left\{
        \begin{aligned}
            c_x&=\frac{\partial z}{\partial x}=\frac{f_x}{(u-u_0)+ze_u},\\
		c_y&=\frac{\partial z}{\partial y}=\frac{f_y}{(v-v_0)+ze_v},
        \end{aligned}
        \right.
    \end{equation}
    where $e_u=\frac{\partial u}{\partial z}$ and $e_v=\frac{\partial v}{\partial z}$ represent the inverse partial derivatives of depth in the $u$ and $v$ directions within the pixel coordinate system, respectively.

    In addition to computing partial derivatives in the horizontal and vertical directions (\textit{i.e.}, the $x$ and $y$ directions) in the camera coordinate system, we extend the computations to include first-order partial derivatives in the diagonal directions (\textit{i.e.}, the $l$ and $t$ directions as depicted in Fig. \ref{frame}). These derivatives are denoted as $c_l$ and $c_t$, respectively, and are calculated as follows:
    \begin{equation}
        \left\{
        \begin{aligned}
            c_l &= \frac{\partial z}{\partial l}=\frac{\sqrt{2}}{2}(\frac{f_x}{(u-u_0)+ze_u}+\frac{f_y}{(v-v_0)+ze_v}),\\
            c_t &= \frac{\partial z}{\partial t}=\frac{\sqrt{2}}{2}(\frac{f_x}{(u-u_0)+ze_u}-\frac{f_y}{(v-v_0)+ze_v}).
        \end{aligned}
        \right.
    \end{equation}
    where the subscripts $l$ and $t$ represent the leading diagonal and trailing diagonal, respectively.

    Within the camera coordinate system, the second-order partial derivatives in the horizontal, vertical, and diagonal directions are subsequently achieved by the second-order partial derivatives of depth in the $u$ and $v$ directions as follows:
    \begin{equation}
        \left\{
            \begin{aligned}
            c_{xx}&=\frac{\partial^2 z}{\partial x^2}=-\frac{f_x^2\alpha_0}{\alpha_1^3},\quad
            c_{yy}=\frac{\partial^2 z}{\partial y^2}=-\frac{f_y^2\beta_0}{\beta_1^3},\\
            c_{ll}&=\frac{\partial^2 z}{\partial l^2}=\frac{\sqrt{2}}{2}(c_{xx}+c_{yy})=-\frac{\sqrt{2}}{2}(\frac{f_x\alpha_0}{\alpha_1^3}+\frac{f_y\beta_0}{\beta_1^3}),\\
            c_{tt}&=\frac{\partial^2 z}{\partial t^2}=\frac{\sqrt{2}}{2}(c_{xx}-c_{yy})=\frac{\sqrt{2}}{2}(\frac{f_y\beta_0}{\beta_1^3}-\frac{f_x\alpha_0}{\alpha_1^3}),
        \end{aligned}
            \right.
            \label{con:comb_diag}
        \end{equation}
        where
        \begin{equation}
            \left\{
            \begin{aligned}
                 \alpha_0& =e_u(2-ze_u^2e_{uu}), \alpha_1=u-u_0+ze_{u}, e_{uu}=\frac{\partial^2 z}{\partial u^2},\\
                 \beta_0& =e_v(2-ze_v^2e_{vv}), \beta_1=v-v_0+ze_{v}, e_{vv}=\frac{\partial^2 z}{\partial v^2}.
            \end{aligned}
            \right.
        \end{equation}
    As shown in Fig. \ref{frame}, the $l$ and $t$ directions in the camera coordinate system respectively correspond to the $g$ and $b$ directions in the pixel coordinate system. We define the depth curvature in each direction in the $3\times3$ local region as $k_u$, $k_v$, $k_g$, and $k_b$, which can be computed as follows:
    \begin{equation}
        \left\{
        \begin{aligned}
            k_u&=\frac{|c_{xx}|}{(1+{c_x}^2)^\frac{3}{2}}=|\frac{f_x^2\alpha_0}{(\alpha_1^2+f_x^2)^\frac{3}{2}}|,\\
            k_v& =\frac{|c_{yy}|}{(1+{c_y}^2)^\frac{3}{2}}=|\frac{f_y^2\beta_0}{(\beta_1^2+f_y^2)^\frac{3}{2}}|,\\
             k_g& =\frac{|c_{ll}|}{(1+{c_l}^2)^\frac{3}{2}}=2|\frac{f_x\alpha_0\beta_1^3+f_y\beta_0\alpha_1^3}{[2\alpha_1^2\beta_1^2+(f_x\beta_1+f_y\alpha_1)^2]^\frac{3}{2}}|,\\
             k_b& =\frac{|c_{tt}|}{(1+{c_t}^2)^\frac{3}{2}}=2|\frac{f_y\beta_0\alpha_1^3-f_x\alpha_0\beta_1^3}{[2\alpha_1^2\beta_1^2+(f_x\beta_1+f_y\alpha_1)^2]^\frac{3}{2}}|.
        \end{aligned}
        \right.
    \end{equation}
    Among the eight neighboring points in the pixel coordinate system, we assign the curvatures of points in the horizontal and vertical directions as $k_{u_1}$, $k_{u_2}$, $k_{v_1}$, and $k_{v_2}$, the curvatures of the two points in the leading diagonal direction as $k_{g_1}$ and $k_{g_2}$, and the curvatures of the two points in the trailing diagonal direction as $k_{b_1}$ and $k_{b_2}$. The maximum curvature $k_{\max}$ at point $\boldsymbol{q}$ on the surface can therefore be obtained as follows:
    \begin{equation}
             k_{\max}=\max\limits_{i=1,2}(k_{u_i}, k_{v_i}, k_{g_i}, k_{b_i}).
    \end{equation}
    Similarly, the mean curvature $k_\text{mean}$ at point $\boldsymbol{q}$ can be computed as follows:
    \begin{equation}
             k_\text{mean}=\sum_{i=1}^2(k_{u_i}+k_{v_i}+k_{g_i}+k_{b_i})/8.
    \end{equation}
    In contrast to the maximum and mean depth curvatures, which concentrate on measuring the curvature of a surface at a given point, normal and Gaussian depth curvatures are used to depict the surface smoothness in a specific direction through the computations of the first and second fundamental forms:
    \begin{equation}
        \begin{split}
             k_\text{normal} = \frac{Le_v^2+2Me_ue_v+Ne_u^2}{Ee_v^2+2Fe_ue_v+Ge_u^2}, \quad k_\text{gauss} = \frac{LN-M^2}{EG-F^2},
        \end{split}
    \end{equation}
    where $E, F,$ and $G$ are the coefficients of the first fundamental form. $L, M,$ and $N$ are the coefficients of the second fundamental form. These terms can be computed as follows:
    \begin{equation}
        \left\{
        \begin{aligned}
             L& =c_{xx}c_{yy}=\frac{f_x^2f_y^2\alpha_0\beta_0}{\alpha_1^3\beta_1^3},\\
             M& =c_{xx}c_{ll}=\frac{\sqrt{2}}{2}(\frac{f_x^3\alpha_0^2}{\alpha_1^6}+\frac{f_x^2f_y\alpha_0\beta_0}{\alpha_1^3\beta_1^3}),\\
             N& =c_{tt}c_{yy}=\frac{\sqrt{2}}{2}(\frac{f_xf_y^2\alpha_0\beta_0}{\alpha_1^3\beta_1^3}-\frac{f_y^3\beta_0^2}{\beta_1^6}),\\
             E& =c_x^2=\frac{f_x^2}{\alpha_1^2},  G=c_y^2=\frac{f_y^2}{\beta_1^2},  F=c_xc_y=\frac{f_xf_y}{\alpha_1\beta_1}.
        \end{aligned}
        \right.
    \end{equation}
    The four depth curvatures are employed to quantify the local plane's smoothness. A higher depth curvature indicates a steeper surface within the point's neighborhood, implying a higher probability of discontinuities. Consequently, the optimum surface normal $\boldsymbol{n}_{\boldsymbol{q}}$ can be determined by considering the depth curvature of the target point $\boldsymbol{q}$ along with its neighboring points. The surface normal estimation for the target point is represented by the surface normal of the neighboring point $\boldsymbol{r}$ with the smallest depth curvature, as follows:
    \begin{equation}
        \begin{split}
            \boldsymbol{r}&=\underset{\boldsymbol{\boldsymbol{r}\in \boldsymbol{P}^{+}}}{\arg\min} \ k_{\boldsymbol{r}},\quad \boldsymbol{n}_{\boldsymbol{q}}=\boldsymbol{n}_{\boldsymbol{r}},
        \end{split}
    \end{equation}
    where we have the flexibility to choose any one of the curvatures $k_{\max}$, $k_{\text{mean}}$, $k_{\text{normal}}$, and $k_{\text{gauss}}$ to represent the depth curvature $k_{\boldsymbol{r}}$. This article demonstrates that the surface normal estimation for the target point is well-suited to the plane in which it is located.

    \subsection{Correlation Coefficient-Based Discontinuity Discrimination}
    \label{sec:corelation}
    To comprehensively identify discontinuities, we not only utilize depth curvature but also incorporate Kendall's \cite{kendall1938new} and Pearson's correlation coefficients \cite{lian2022research}. We perform a mathematical analysis of various combinations of depth values for both the target point and its neighboring points. In accordance with the provided depth map, we define point pairs $(z_{\boldsymbol{q}}, z_{\boldsymbol{p}_i})$ for the computation of Kendall's correlation coefficient within a local $3\times3$ region. Subsequently, we evaluate the consistency of these pairs. Following the calculation principles of Kendall's correlation coefficient, consistent combinations are those in which the depth of the target point is less than the depth of the neighboring points, resulting in a total of $\mu_c$ group pairs. Inconsistent combinations, on the other hand, occur when the depth of the target point exceeds that of the neighboring points, resulting in a total of $\mu_d$ combinations. It is important to note that if the target point and neighboring points within a point pair share identical depth values, neither consistent nor inconsistent combinations are considered.

    We define Kendall's correlation coefficient for discriminating discontinuous regions as $\tau$, which is determined by the ratio of consistent to inconsistent pairs. This coefficient ranges from 0 to 1, with larger values indicating a higher probability that the points belong to the same continuous plane. Consequently, as the coefficient increases, the surface normal of the target point becomes more aligned with the surface normal of neighboring points. The final surface normal $\boldsymbol{n}_{\boldsymbol{q}}$ for the target point is determined by selecting the neighboring point $\boldsymbol{r}$ with the maximum Kendall's correlation coefficient $\tau_{\boldsymbol{r}}$ as follows:
    \begin{equation}
        \left\{
        \begin{aligned}
             \tau_{\boldsymbol{r}} & = |\frac{\mu_c-\mu_d}{\mu_c+\mu_d}|, \boldsymbol{r}\in \boldsymbol{P}^{+},\\
             \boldsymbol{r}& =\underset{\boldsymbol{\boldsymbol{r}\in \boldsymbol{P}^{+}}}{\arg\max} \ \tau_{\boldsymbol{r}},\quad \boldsymbol{n}_{\boldsymbol{q}}=\boldsymbol{n}_{\boldsymbol{r}}.
        \end{aligned}
        \right.
    \end{equation}
    To calculate Pearson's correlation coefficient $\varepsilon_{\boldsymbol{r}}$ for point $\boldsymbol{r}$, we begin by constructing a new sequence composed of the depths of neighboring points, denoted as $\boldsymbol{\psi} =(z_{\boldsymbol{p}_1},..., z_{\boldsymbol{p}_8})$. We then create a sequence $\boldsymbol{\psi}^+=(z_{\boldsymbol{q}}, z_{\boldsymbol{p}_1},..., z_{\boldsymbol{p}_8})$. The Pearson's correlation coefficient can be calculated as follows:
    \begin{equation}
        \left\{
        \begin{aligned}
             \varepsilon_{\boldsymbol{r}}& = \frac{|(\boldsymbol{\psi}-\bar{\boldsymbol{\psi}})(\boldsymbol{\psi}^+-\bar{\boldsymbol{\psi}}^+)|}
             {\sqrt{||\boldsymbol{\psi}-\bar{\boldsymbol{\psi}}||^2||\boldsymbol{\psi}^+-\bar{\boldsymbol{\psi}}^+||^2}},\boldsymbol{r}\in \boldsymbol{P}^{+},\\
             \boldsymbol{r}& =\underset{\boldsymbol{\boldsymbol{r}\in \boldsymbol{P}^{+}}}{\arg\max} \ \varepsilon_{\boldsymbol{r}},\quad \boldsymbol{n}_{\boldsymbol{q}}=\boldsymbol{n}_{\boldsymbol{r}},
        \end{aligned}
        \right.
    \end{equation}
    where the coefficient $\varepsilon_{\boldsymbol{r}}$ ranges from 0 to 1, and $\boldsymbol{n}_{\boldsymbol{q}}$ represents the optimized surface normal via discontinuity discrimination.

		\begin{figure*}[t!]
		\centering
		\includegraphics[width=0.9999\linewidth]{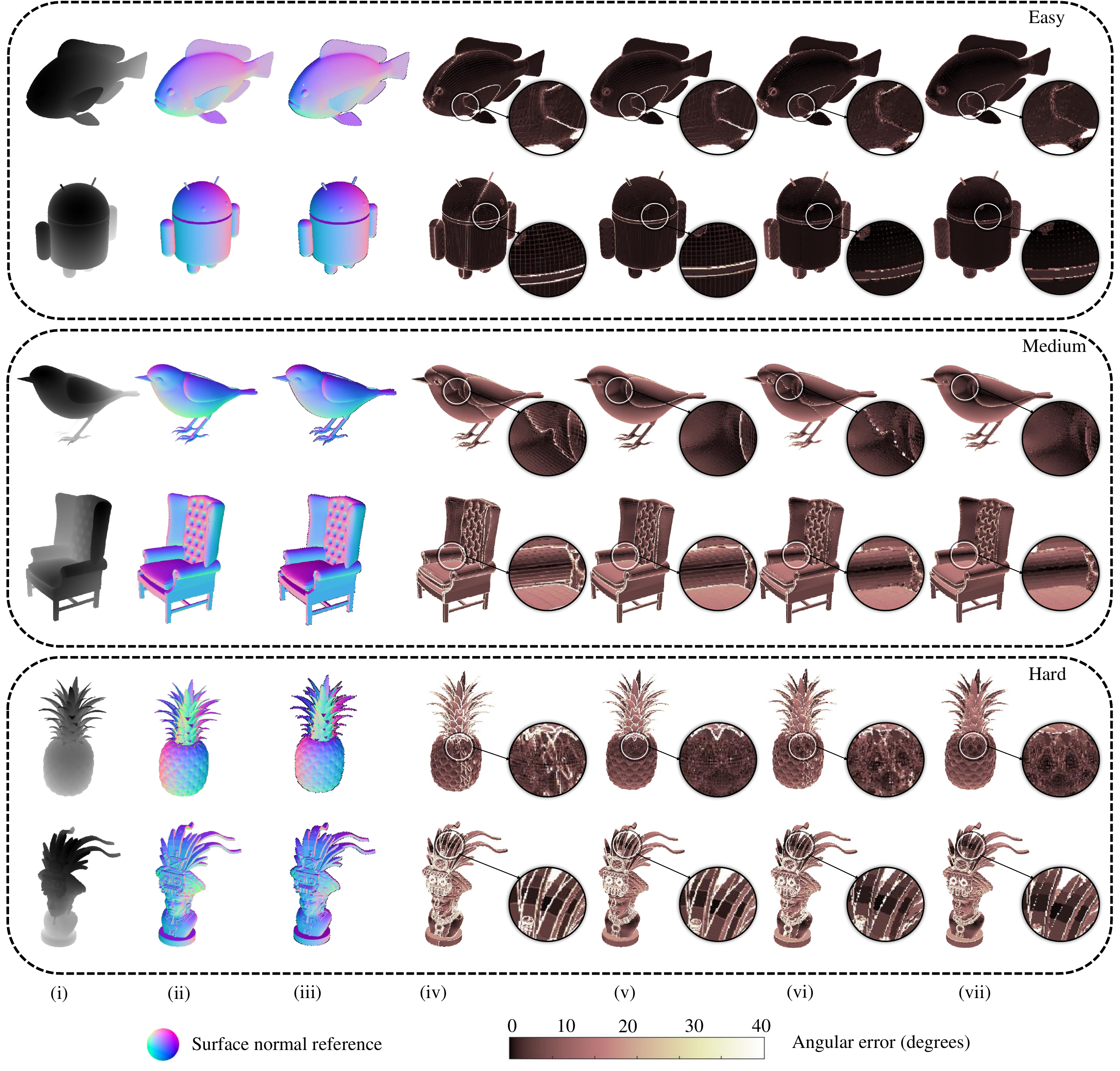}
		\caption{Examples of surface normal estimation results on the 3F2N-Easy, 3F2N-Medium and 3F2N-Hard datasets. (i) and (ii) show depth and surface normal ground truth, respectively. (iii) shows the surface normal estimation results achieved by 3F2N+. (iv)-(vii) show angular error maps obtained by AngleWeighted, 3F2N, AngleWeighted+, and 3F2N+, respectively.}
		\label{3f2ndataset}
	\end{figure*}

        \begin{figure*}[t!]
		\centering
		\includegraphics[width=0.9999\linewidth]{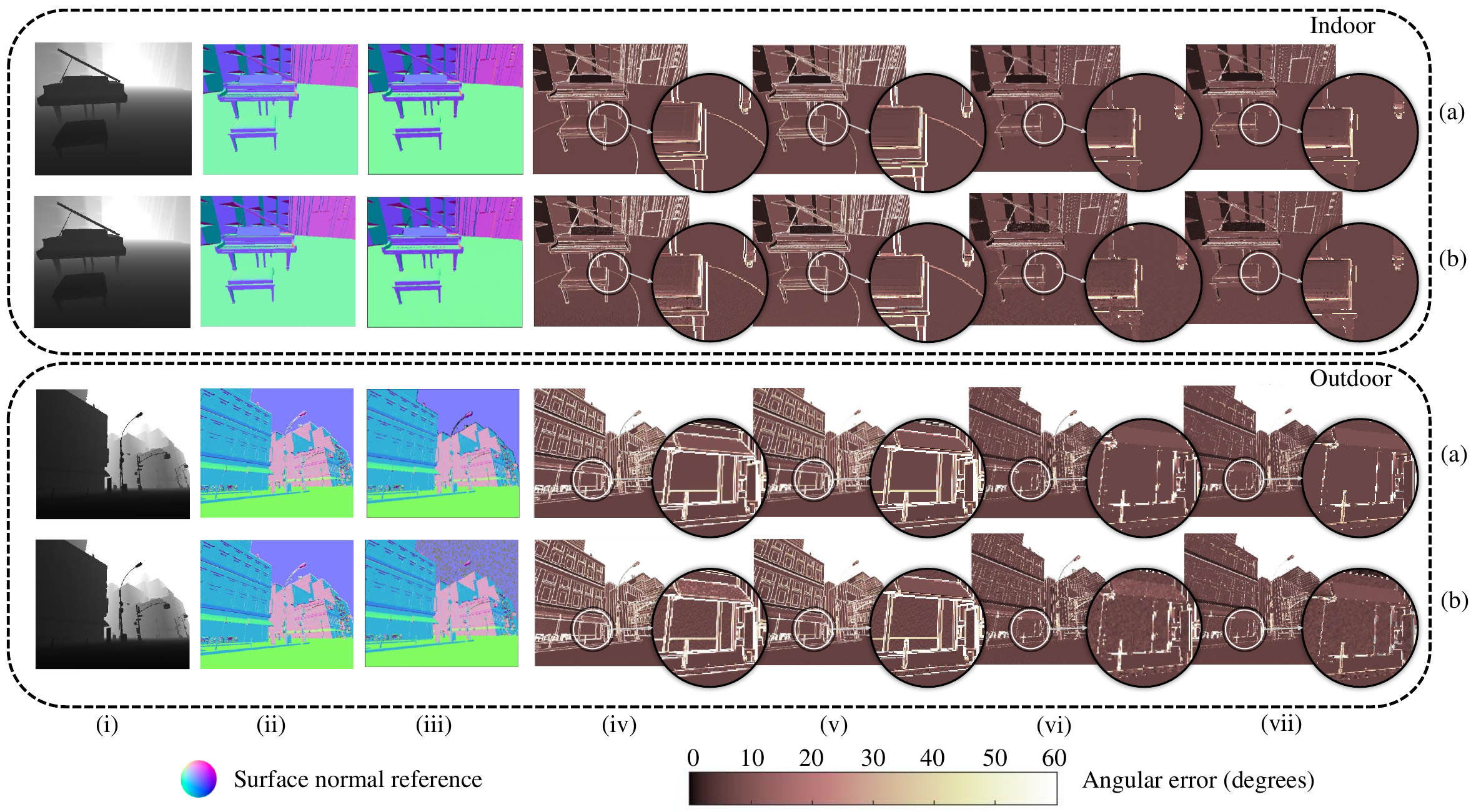}
		\caption{Examples of surface normal estimation results on the SSN dataset. (i) and (ii) show depth and surface normal ground truth, respectively. (iii) shows the surface normal estimation results achieved by 3F2N+. (iv)-(vii) show angular error maps obtained by AngleWeighted, 3F2N, AngleWeighted+, and 3F2N+, respectively. (a) and (b) show the results with respect to clean and noisy inputs, respectively.}
		\label{ourdataset}
	\end{figure*}

    \subsection{Discontinuity Discrimination Module}
    \label{sec:ddm}
    We introduce a CRF-based surface normal optimization method that combines the two discontinuity discrimination strategies, respectively discussed in Sect. \ref{sec:curvation} and Sect. \ref{sec:corelation}, to further minimize the surface normal estimation error in discontinuous regions.
    
    For a given depth image, we first define an undirected graph $\boldsymbol{G}=(\boldsymbol{Q},\boldsymbol{E})$, where the nodes $\boldsymbol{Q}$ denote pixels in the depth image, while the edges $\boldsymbol{E}$ denote the correlations between adjacent pixels. We then define $P(\boldsymbol{N})$ as a conditional probability distribution that characterizes the surface normal $\boldsymbol{n}_{\boldsymbol{q}}$ within the surface normal field $\boldsymbol{N}$ as follows:
    \begin{equation}
        \begin{split}
             P(\boldsymbol{N}) = \frac{1}{\Theta(\boldsymbol{N})} \exp \left(\sum_{\boldsymbol{q} \in \boldsymbol{Q}} \sigma f(\boldsymbol{q})+\sum_{(\boldsymbol{p}_i,\boldsymbol{q})\in \boldsymbol{E}} \lambda_i g(\boldsymbol{p}_i,\boldsymbol{q})\right),
        \end{split}
    \end{equation}
    where $\Theta(\boldsymbol{N})$ is the partition function that normalizes the conditional probability distribution, $\sigma$ and $\lambda_i$ represent weights of $f(\cdot)$ and $g(\cdot)$, respectively.
    \begin{equation}
        \begin{split}
            f(\boldsymbol{q})=\|\boldsymbol{n}_{\boldsymbol{q}}-\hat{\boldsymbol{n}}_{\boldsymbol{q}}\|
        \end{split}
    \end{equation}
    is the node feature function used to compute the cost of the surface normal of the point $\boldsymbol{q}$, thereby finding the best estimation of the surface normal $\hat{\boldsymbol{n}}_{\boldsymbol{q}}$. 
    \begin{equation}
        \begin{split}
            g(\boldsymbol{p}_i,\boldsymbol{q})=k_{\boldsymbol{q}}\sum\limits_{i}(1-\tau_{\boldsymbol{p}_i}\vee\varepsilon_{\boldsymbol{p}_i})\|\boldsymbol{n}_{\boldsymbol{p}_i}-\hat{\boldsymbol{n}}_{\boldsymbol{q}}\|
        \end{split}
        \label{equ:g}
    \end{equation}
    denotes the edge feature function, which is used to penalize the surface normal discontinuity among pairs of neighboring pixels, where $k_{\boldsymbol{q}}$ is the depth curvature obtained in Sect. \ref{sec:curvation}. (\ref{equ:g}) illustrates that when neighboring points tend to align with a plane, it is more probable for the distribution of surface normals to exhibit similarity. As a result, this leads to optimized surface normals that are more aligned with the underlying surfaces.
    
\begin{table}[t!]
    \settablefont
    \begin{center}
    \settablefont
    \caption{Comparison of $e_\text{A}$ w.r.t. different central tendency measurements, gradient filters, and depth curvatures.}
    \setlength\arrayrulewidth{0.5pt}
    \begin{tabular}{l|l|l|rrr}
        \hline
        \multicolumn{1}{l|}{\multirow{1}*{Central}}
        & \multicolumn{1}{l|}{\multirow{2}*{Gradient}}
        & \multicolumn{1}{l|}{\multirow{2}*{Depth}}
        &\multicolumn{3}{c}{$e_\text{A}$ (degrees)}\\
        \cline{4-6}
        Tendency&\multicolumn{1}{l|}{\multirow{2}*{Filter}}&\multicolumn{1}{l|}{\multirow{2}*{Curvature}}
        &\multicolumn{1}{c}{\multirow{2}*{Easy}} &\multicolumn{1}{c}{\multirow{2}*{Medium}} & \multicolumn{1}{c}{\multirow{2}*{Hard}}\\
        Measurement&  &   &   &   &  \\
        
        \hline
        \multicolumn{1}{l|}{\multirow{8}*{Median}}
        &\multicolumn{1}{l|}{\multirow{4}*{FD}}
        &Normal
        &1.214
        &5.321
        &11.648
        \\
        
        &
        &Gaussian
        &1.216
        &5.155
        &11.714
        \\
        
        &
        &Maximum
        &0.914
        &4.791
        &9.976
        \\
        
        &
        &\cellcolor{LightGray}Mean
        &\cellcolor{LightGray}\textbf{0.905}
        &\cellcolor{LightGray}\textbf{4.776}
        &\cellcolor{LightGray}\textbf{9.959}
        \\

        \cline{2-6}
        
        &\multicolumn{1}{l|}{\multirow{4}*{Roberts}}
        &Normal
        &1.343
        &5.589
        &11.776
        \\
        
        &
        &Gaussian
        &1.509
        &5.698
        &12.197
        \\
        
        &
        &Maximum
        &1.087
        &5.161
        &11.186
        \\
        
        &
        &\cellcolor{LightGray}Mean
        &\cellcolor{LightGray}1.036
        &\cellcolor{LightGray}5.063
        &\cellcolor{LightGray}10.939
        \\

        \cline{1-6}
        \multicolumn{1}{l|}{\multirow{8}*{Trimean}}
        &\multicolumn{1}{l|}{\multirow{4}*{FD}}
        &Normal
        &1.228
        &5.352
        &11.582
        \\
        
        &
        &Gaussian
        &1.229
        &5.189
        &11.342
        \\
        
        &
        &Maximum
        &0.931
        &4.835
        &10.013
        \\
        
        &
        &\cellcolor{LightGray}Mean
        &\cellcolor{LightGray}0.922
        &\cellcolor{LightGray}4.820
        &\cellcolor{LightGray}9.997
        \\

        \cline{2-6}
        
        &\multicolumn{1}{l|}{\multirow{4}*{Roberts}}
        &Normal
        &1.371
        &5.640
        &11.873
        \\
        
        &
        &Gaussian
        &1.576
        &5.808
        &12.447
        \\
        
        &
        &Maximum
        &1.087
        &5.161
        &11.186
        \\
        
        &
        &\cellcolor{LightGray}Mean
        &\cellcolor{LightGray}1.079
        &\cellcolor{LightGray}5.152
        &\cellcolor{LightGray}11.155
        \\

        \cline{1-6}
        \multicolumn{1}{l|}{\multirow{8}*{Trimmean}}
        &\multicolumn{1}{l|}{\multirow{4}*{FD}}
        &Normal
        &1.212
        &5.305
        &11.830
        \\
        
        &
        &Gaussian
        &1.518
        &5.160
        &11.391
        \\
        
        &
        &Maximum
        &0.939
        &4.853
        &10.035
        \\
        
        &
        &\cellcolor{LightGray}Mean
        &\cellcolor{LightGray}0.929
        &\cellcolor{LightGray}4.839
        &\cellcolor{LightGray}10.019
        \\    
        
        \cline{2-6}
        
        &\multicolumn{1}{l|}{\multirow{4}*{Roberts}}
        &Normal
        &1.704
        &5.581
        &12.151
        \\
        
        &
        &Gaussian
        &2.071
        &5.791
        &12.763
        \\
        
        &
        &Maximum
        &1.102
        &5.195
        &11.245
        \\
        
        &
        &\cellcolor{LightGray}Mean
        &\cellcolor{LightGray}1.095
        &\cellcolor{LightGray}5.187
        &\cellcolor{LightGray}11.219
        \\
        \hline        
    \end{tabular}
    \label{curvature_compare_fd}
    \end{center}
\end{table}

\section{Experiments}
\label{sec.exp}

\subsection{Datasets and Evaluation Metrics}
In our previous study \cite{fan2020sne}, we introduced a large-scale, noise-less, synthetic dataset consisting of three subsets: 3F2N-Easy, 3F2N-Medium, and 3F2N-Hard. In this article, we create SSN dataset using 20 3D mesh models, including depth images with and without the addition of random Gaussian noise. Our dataset is divided into two subsets: indoor and outdoor scenarios. For each scenario model, we capture 500 different views using an Intel Core i7-12700H CPU. Z-Buffer rendering \cite{greene1993hierarchical} is employed to generate depth images with a resolution of $512\times424$ pixels. All scenario models are represented by meshes composed of triangles, the surface normals of which serve as the ground truth. The camera parameters used in our dataset are identical to those of the Kinect v2 sensor.

The accuracy of SNEs is quantified using the average angular error:
\begin{equation}
	\begin{split}
		e_\text{A}=\frac{1}{m}\sum_{k=1}^{m}\cos^{-1}\frac{\langle \boldsymbol{n}_k, \hat{\boldsymbol{n}}_k \rangle}{||\boldsymbol{n}_k||_2||\hat{\boldsymbol{n}}_k||_2},
	\end{split}\label{con:eq41}
\end{equation}
where $m$ denotes the number of observed points, while $\boldsymbol{n}_k$ and $\hat{\boldsymbol{n}}_k$ denote the ground truth and estimated surface normals, respectively.
\subsection{Performance Evaluation of Depth Curvature-Based Discontinuity Discrimination Approaches}
\label{sec:exp_curv}
The evaluation results of the four depth curvature-based discontinuity discrimination methods are presented in Table \ref{curvature_compare_fd}. As expected, the results obtained using mean and maximum curvatures demonstrate higher accuracy compared to those obtained using normal and Gaussian curvatures. The mean curvature stands out with the most superior performance, making it the most robust choice in our following experiments. Additionally, 3F2N+ using the finite difference (FD) filter, median filter, and mean curvature, achieves an inference speed of 84 FPS.

\begin{figure*}[t!]
		\centering
		\includegraphics[width=0.999\linewidth]{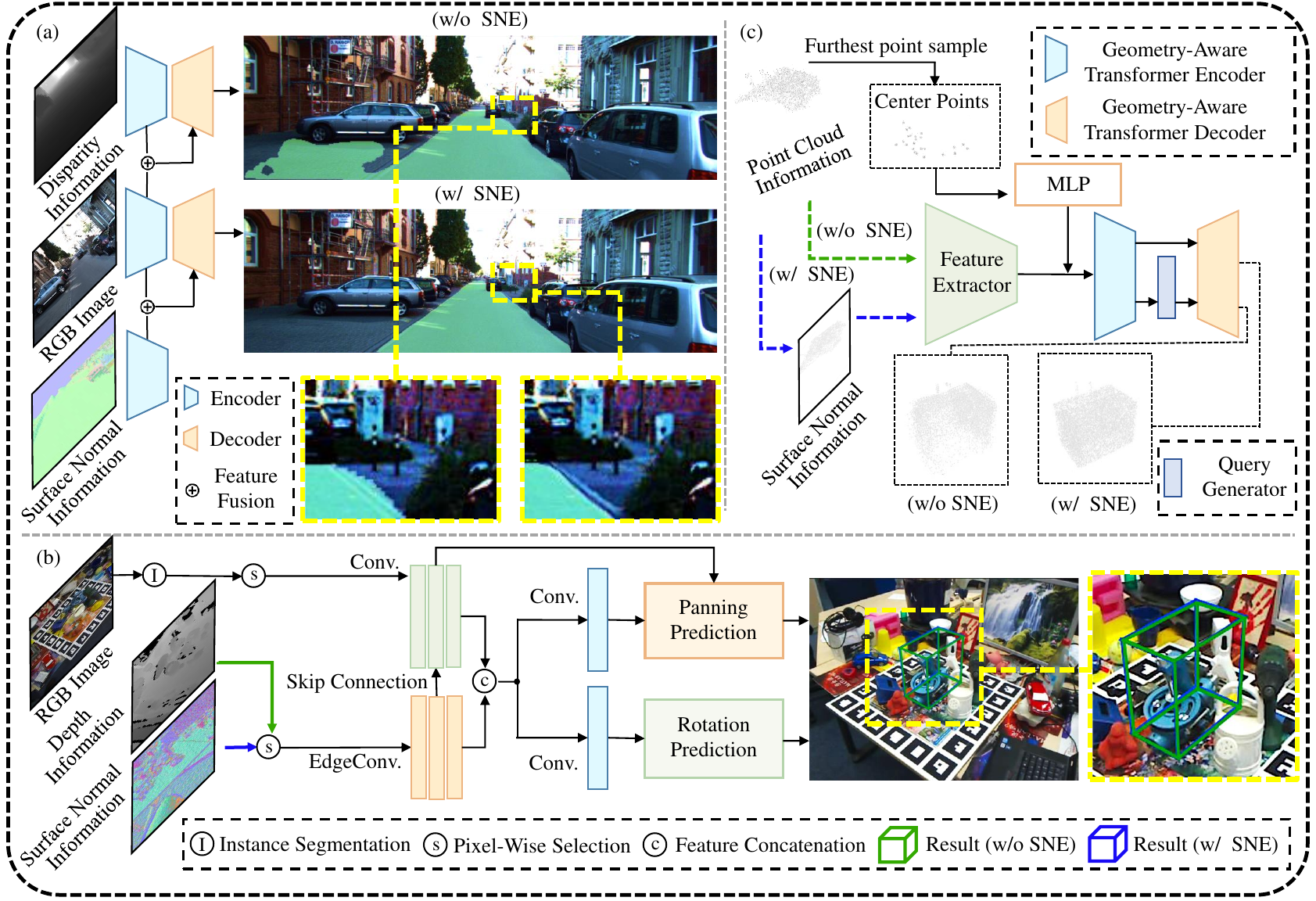}
		\caption{Robot perception tasks using 3F2N+: (a) data-fusion freespace detection \cite{fan2020sne}; (b) 6D object pose estimation \cite{liu2023bdr6d}; (c) point cloud completion \cite{huang2020pf}.}
		\label{cover}
\end{figure*}
\begin{table}[t!]
    \settablefont
    \begin{center}
    \setlength{\tabcolsep}{12pt}
    \caption{$e_\text{A}$ comparisons between 3F2N and 3F2N+ w/ correlation coefficient-based discontinuity discrimination}.
    \setlength\arrayrulewidth{0.8pt}
    \begin{tabular}{l|ccc}
        \cline{1-4}
        \multicolumn{1}{l|}{\multirow{2}{*}{Method}} 
        & \multicolumn{3}{c}{$e_\text{A}$ (degrees)} \\
        \cline{2-4}

        &Easy
        &Medium
        &Hard
        \\
        \hline
        
        3F2N
        &1.657
        &5.685
        &15.313
        \\
        
        3F2N+ w/ Pearson
        &\textbf{1.320}
        &5.645
        &13.122
        \\

        3F2N+ w/ Kendall
        &1.352
        &\textbf{5.360}
        &\textbf{12.584}
        \\
        \hline
        
    \end{tabular}
    \label{tab:cor}
    \end{center}
\end{table}

\begin{table*}[t!]
\settablefont
    \centering
    \setlength{\tabcolsep}{9pt}
    \caption{$e_\text{A}$ and the runtime comparisons among SoTA SNEs on the 3F2N and SSN datasets.}
    \begin{tabular}{l|c|p{2cm}p{2cm}p{2cm}|p{2cm}p{2cm}p{2cm}p{2cm}}
        \cline{1-9}
        \multicolumn{1}{l|}{\multirow{3}{*}{Method}} 
        & \multicolumn{1}{l|}{\multirow{3}{*}{Runtime (ms)}}
        & \multicolumn{7}{c}{$e_\text{A}$ (degrees)} \\
        \cline{3-9}   
        & 
        & \multicolumn{3}{c|}{3F2N dataset}
        & \multicolumn{4}{c}{SSN dataset}\\      
        \cline{3-9}
        &
        &\multicolumn{1}{c}{Easy}&    \multicolumn{1}{c}{Medium} & \multicolumn{1}{c|}{Hard}
        &\multicolumn{1}{c}{Clean-Indoor}& \multicolumn{1}{c}{Noisy-Indoor} &\multicolumn{1}{c}{Clean-Outdoor}&\multicolumn{1}{c}{Noisy-Outdoor}\\ 
        \cline{1-9}
        \multirow{1}{*}{PlaneSVD \cite{p6}} 
        &336.63
        &\multicolumn{1}{r}{\multirow{1}{*}{2.07}}
        &\multicolumn{1}{c}{\multirow{1}{*}{6.07}}
        &\multicolumn{1}{r|}{\multirow{1}{*}{17.59}}
        &\multicolumn{1}{r}{\multirow{1}{*}{11.61}}
        &\multicolumn{1}{r}{\multirow{1}{*}{13.05}}
        &\multicolumn{1}{r}{\multirow{1}{*}{13.69}}
        &\multicolumn{1}{r}{\multirow{1}{*}{14.43}}\\        
        \multirow{1}{*}{PlanePCA \cite{klasing2009realtime}} 
        &540.29
        &\multicolumn{1}{r}{\multirow{1}{*}{2.07}}
        &\multicolumn{1}{c}{\multirow{1}{*}{6.07}}
        &\multicolumn{1}{r|}{\multirow{1}{*}{17.59}}
        &\multicolumn{1}{r}{\multirow{1}{*}{12.39}}
        &\multicolumn{1}{r}{\multirow{1}{*}{13.05}}
        &\multicolumn{1}{r}{\multirow{1}{*}{12.91}}
        &\multicolumn{1}{r}{\multirow{1}{*}{14.42}}\\  
        \multirow{1}{*}{VectorSVD \cite{klasing2009comparison}} 
        &481.58
        &\multicolumn{1}{r}{\multirow{1}{*}{2.13}}
        &\multicolumn{1}{c}{\multirow{1}{*}{6.27}}
        &\multicolumn{1}{r|}{\multirow{1}{*}{18.01}}
        &\multicolumn{1}{r}{\multirow{1}{*}{12.72}}
        &\multicolumn{1}{r}{\multirow{1}{*}{13.61}}
        &\multicolumn{1}{r}{\multirow{1}{*}{13.43}}
        &\multicolumn{1}{r}{\multirow{1}{*}{16.15}}\\    
        \multirow{1}{*}{AreaWeighted \cite{jin2005comparison}} 
        &933.93
        &\multicolumn{1}{r}{\multirow{1}{*}{2.20}}
        &\multicolumn{1}{c}{\multirow{1}{*}{6.27}}
        &\multicolumn{1}{r|}{\multirow{1}{*}{17.03}}
        &\multicolumn{1}{r}{\multirow{1}{*}{11.94}}
        &\multicolumn{1}{r}{\multirow{1}{*}{12.59}}
        &\multicolumn{1}{r}{\multirow{1}{*}{12.47}}
        &\multicolumn{1}{r}{\multirow{1}{*}{14.09}}\\        
        \multirow{1}{*}{AngleWeighted \cite{jin2005comparison}} 
        &883.17
        &\multicolumn{1}{r}{\multirow{1}{*}{1.79}}
        &\multicolumn{1}{c}{\multirow{1}{*}{5.67}}
        &\multicolumn{1}{r|}{\multirow{1}{*}{13.26}}
        &\multicolumn{1}{r}{\multirow{1}{*}{9.51}}
        &\multicolumn{1}{r}{\multirow{1}{*}{10.17}}
        &\multicolumn{1}{r}{\multirow{1}{*}{10.37}}
        &\multicolumn{1}{r}{\multirow{1}{*}{12.36}}\\
 
        \multirow{1}{*}{FALS \cite{badino2011fast}} 
        &3.51
        &\multicolumn{1}{r}{\multirow{1}{*}{2.26}}
        &\multicolumn{1}{c}{\multirow{1}{*}{6.14}}
        &\multicolumn{1}{r|}{\multirow{1}{*}{17.34}}
        &\multicolumn{1}{r}{\multirow{1}{*}{12.11}}
        &\multicolumn{1}{r}{\multirow{1}{*}{12.55}}
        &\multicolumn{1}{r}{\multirow{1}{*}{12.47}}
        &\multicolumn{1}{r}{\multirow{1}{*}{13.78}}\\
        
        \multirow{1}{*}{SRI \cite{badino2011fast}} 
        &10.41
        &\multicolumn{1}{r}{\multirow{1}{*}{2.64}}
        &\multicolumn{1}{c}{\multirow{1}{*}{6.71}}
        &\multicolumn{1}{r|}{\multirow{1}{*}{19.61}}
        &\multicolumn{1}{r}{\multirow{1}{*}{13.82}}
        &\multicolumn{1}{r}{\multirow{1}{*}{14.02}}
        &\multicolumn{1}{r}{\multirow{1}{*}{13.73}}
        &\multicolumn{1}{r}{\multirow{1}{*}{14.79}}\\
        
        \multirow{1}{*}{LINE-MOD \cite{hinterstoisser2011gradient}} 
        &5.50
        &\multicolumn{1}{r}{\multirow{1}{*}{6.53}}
        &\multicolumn{1}{c}{\multirow{1}{*}{9.94}}
        &\multicolumn{1}{r|}{\multirow{1}{*}{31.45}}
        &\multicolumn{1}{r}{\multirow{1}{*}{20.55}}
        &\multicolumn{1}{r}{\multirow{1}{*}{20.58}}
        &\multicolumn{1}{r}{\multirow{1}{*}{18.51}}
        &\multicolumn{1}{r}{\multirow{1}{*}{18.64}}\\
        
        \multirow{1}{*}{SNE-RoadSeg \cite{fan2020sne}} 
        &6.77
        &\multicolumn{1}{r}{\multirow{1}{*}{2.04}}
        &\multicolumn{1}{c}{\multirow{1}{*}{6.28}}
        &\multicolumn{1}{r|}{\multirow{1}{*}{16.37}}
        &\multicolumn{1}{r}{\multirow{1}{*}{9.10}}
        &\multicolumn{1}{r}{\multirow{1}{*}{12.68}}
        &\multicolumn{1}{r}{\multirow{1}{*}{11.13}}
        &\multicolumn{1}{r}{\multirow{1}{*}{16.14}}\\
        
        \multirow{1}{*}{3F2N (mean) \cite{fan2021three}} 
        &\textbf{3.18}
        &\multicolumn{1}{r}{\multirow{1}{*}{2.14}}
        &\multicolumn{1}{c}{\multirow{1}{*}{6.66}}
        &\multicolumn{1}{r|}{\multirow{1}{*}{15.30}}
        &\multicolumn{1}{r}{\multirow{1}{*}{11.07}}
        &\multicolumn{1}{r}{\multirow{1}{*}{16.11}}
        &\multicolumn{1}{r}{\multirow{1}{*}{12.15}}
        &\multicolumn{1}{r}{\multirow{1}{*}{18.16}}\\
        
        \multirow{1}{*}{3F2N (median) \cite{fan2021three}} 
        &9.38
        &\multicolumn{1}{r}{\multirow{1}{*}{1.66}}
        &\multicolumn{1}{c}{\multirow{1}{*}{5.69}}
        &\multicolumn{1}{r|}{\multirow{1}{*}{15.31}}
        &\multicolumn{1}{r}{\multirow{1}{*}{11.56}}
        &\multicolumn{1}{r}{\multirow{1}{*}{13.08}}
        &\multicolumn{1}{r}{\multirow{1}{*}{12.28}}
        &\multicolumn{1}{r}{\multirow{1}{*}{15.35}}\\
        
        \multirow{1}{*}{\cellcolor{LightGray}3F2N+ (ours)} 
        &\cellcolor{LightGray}22.96
        & \multicolumn{1}{r}{\multirow{1}{*}{\cellcolor{LightGray}\textbf{0.93}}}
        &\multicolumn{1}{c}{\multirow{1}{*}{\cellcolor{LightGray}\textbf{4.07}}}
        & \multicolumn{1}{r|}{\multirow{1}{*}{\cellcolor{LightGray}\textbf{9.98}}}
        &\multicolumn{1}{r}{\multirow{1}{*}{\cellcolor{LightGray}\textbf{7.85}}}
        &\multicolumn{1}{r}{\multirow{1}{*}{\cellcolor{LightGray}\textbf{9.25}}}
        &\multicolumn{1}{r}{\multirow{1}{*}{\cellcolor{LightGray}\textbf{8.95}}}
        &\multicolumn{1}{r}{\multirow{1}{*}{\cellcolor{LightGray}\textbf{11.98}}}\\
        \cline{1-9}
    \end{tabular}
    \label{3F2N+}
\end{table*}

\subsection{Performance Evaluation of Correlation Coefficient-Based Discontinuity Discrimination Approaches}
\label{sec:exp_cor}
The quantitative results of two correlation coefficient-based discontinuity discrimination methods combined with 3F2N+ (denoted as 3F2N+ w/ Pearson and 3F2N+ w/ Kendall) are presented in Table \ref{tab:cor}. The $e_\text{A}$ scores for 3F2N+ w/ Pearson and 3F2N+ w/ Kendall are 11.33\% and 14.82\% lower than those achieved using 3F2N on the 3F2N datasets, respectively, with 3F2N+ w/ Kendall showing slightly better performance. Additionally, the inference speeds of 3F2N+ w/ Pearson and 3F2N+ w/ Kendall reach 58 FPS and 68 FPS, respectively.

\begin{figure}[t!]
    \centering
    \includegraphics[width=1\linewidth]{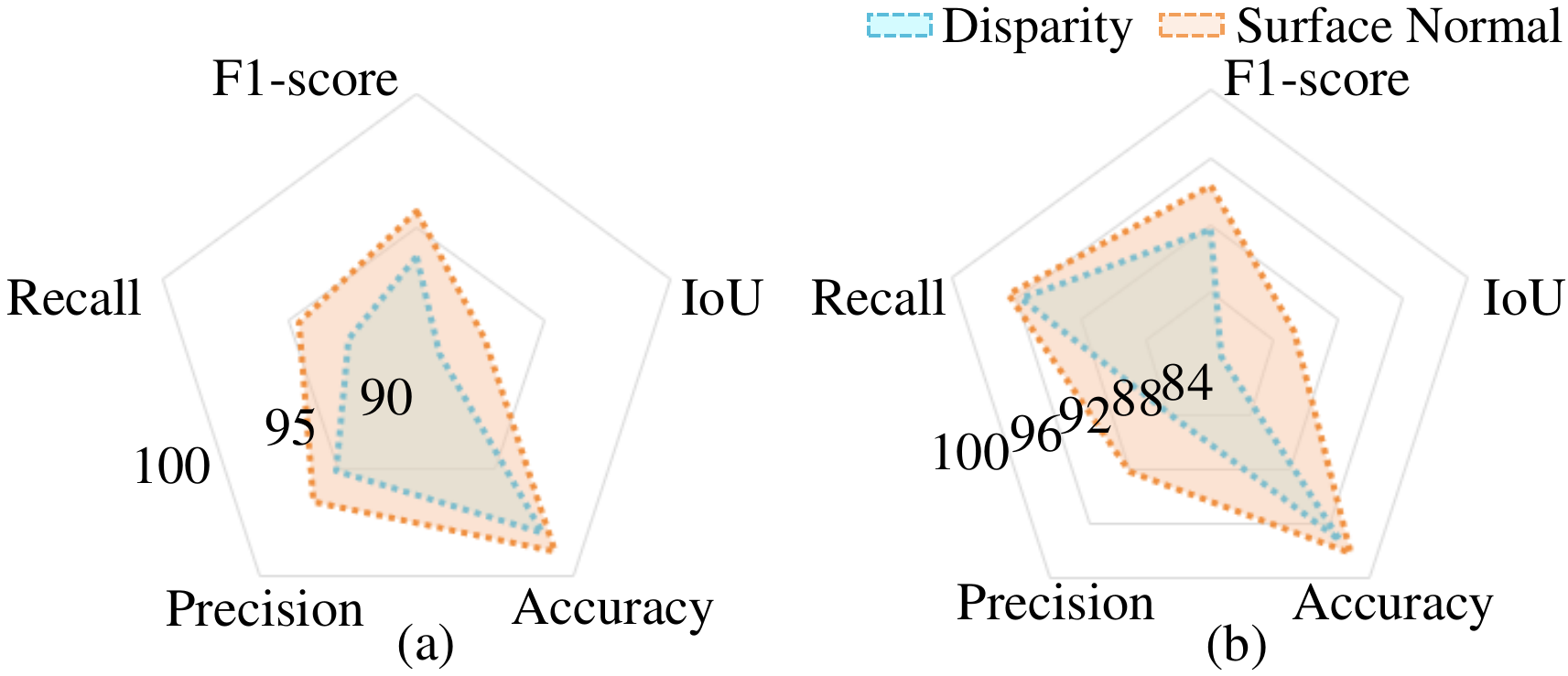}
    \caption{Performance comparison (\%) with respect to different input data for data-fusion freespace detection: (a) MFNet result; (b) FuseNet result.}
    \label{fusion}
\end{figure}

\begin{figure}[t!]
    \centering
    \includegraphics[width=1\linewidth]{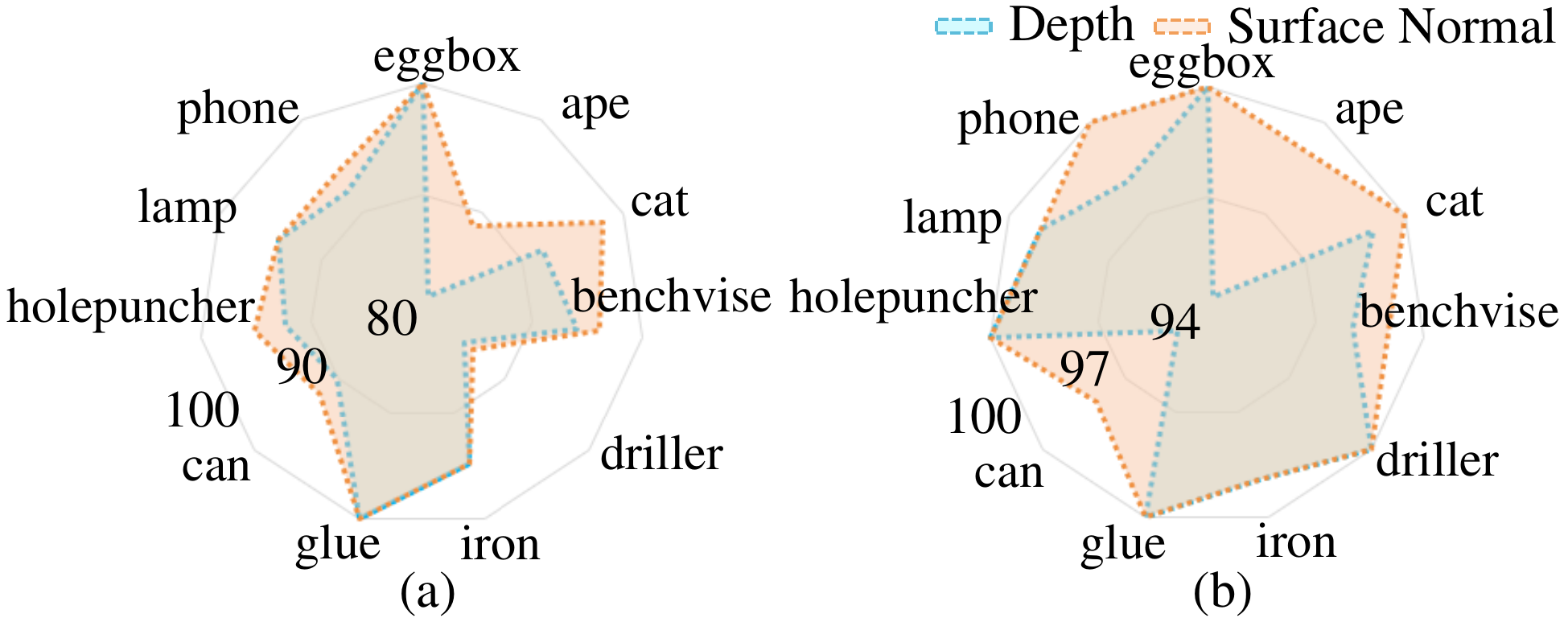}
    \caption{Performance comparison (\%) between two 6D object pose estimation methods: (a) PoseNet; (b) GCN.}
    \label{6dpose}
\end{figure}
\subsection{Performance Evaluation of Discontinuity Discrimination Module}
\label{sec:exp_ddm}
3F2N+ integrates the optimal combination (FD filter, median filter, and mean curvature) obtained in Sect. \ref{sec:exp_curv} along with Kendall's correlation coefficient method mentioned in Sect. \ref{sec:exp_cor} into the DDM. The quantitative results on the 3F2N dataset are presented in Table \ref{3F2N+}. Compared to 3F2N, 3F2N+ achieves a reduction in $e_\text{A}$ by around 44.17\%, 28.50\%, and 34.84\% on the 3F2N-Easy, 3F2N-Medium, and 3F2N-Hard datasets, respectively. Additionally, we employ AngleWeighted combined with DDM (denoted as AngleWeighted+) to further validate the efficacy of this module. The qualitative results shown in Figs. \ref{3f2ndataset} and \ref{ourdataset} demonstrate improved surface normal estimation performance near or on discontinuities with the incorporation of DDM.

\subsection{Performance Evaluation on the 3F2N and SSN Datasets}
The average angular error and the runtime comparisons between SoTA SNEs mentioned in Sect. \ref{sec:rw} and 3F2N+ are presented in Table \ref{3F2N+}. The $e_\text{A}$ scores obtained by 3F2N+ are 0.93$^\circ$, 4.07$^\circ$, and 9.98$^\circ$ on the 3F2N-Easy, 3F2N-Medium, and 3F2N-Hard datasets, respectively. Despite achieving SoTA accuracy, 3F2N+ attains an inference speed of 44 FPS.

In addition, the quantitative and qualitative results of 3F2N+ on the SSN dataset are shown in Table \ref{3F2N+} and Fig. \ref{ourdataset}, respectively, and our proposed method obtains the lowest $e_\text{A}$ scores, which demonstrates the robustness of 3F2N+ on noisy data. All results indicate that 3F2N+ greatly improves the accuracy of surface normal estimation while still ensuring real-time performance.
\section{Application of 3F2N+ in other Robot Perception Tasks}
\label{sec.app}
In this section, we incorporate our proposed 3F2N+ into three robot perception tasks \cite{fan2023autonomous}: (1) data-fusion freespace detection, (2) 6D object pose estimation, and (3) point cloud completion, to further demonstrate its applicability.

First, we train two data-fusion networks: MFNet \cite{ha2017mfnet} and FuseNet \cite{hazirbas2017fusenet}, on the KITTI Road dataset \cite{menze2015object}. For each of these networks, we use two different inputs: RGB+Disparity and RGB+Normal. The results of our experiments are presented in Figs. \ref{cover}(a) and \ref{fusion}. Notably, when we incorporate surface normal information as the input, we observe significant improvements in performance metrics. Specifically, the intersection over union (IoU) metric shows an increase of up to 6\%, and the F1-score demonstrates an increase of up to 3\%.  These findings align with the conclusions drawn in our prior work \cite{yang2023semantic}, providing further empirical evidence to support the effectiveness of the SNE used in freespace detection algorithms.

Figs. \ref{cover}(b) and \ref{6dpose}, as well as Table \ref{6d_acc_dis} present qualitative and quantitative results of 6D object pose estimation on the LineMOD dataset \cite{hinterstoisser2011multimodal} with inputs of RGB+Depth data and RGB+Normal data. Table \ref{6d_acc_dis} demonstrates that the accuracy is improved by 1.1\% and 1.6\%, respectively, and the average 3D distance (ADD) \cite{he2021ffb6d} decreases by 0.049 mm and 0.670 mm, respectively, when using PoseNet \cite{kendall2015posenet} and graph convolutional network (GCN) \cite{kipf2017semisupervised}. These results strongly suggest that our proposed 3F2N+ can serve as an effective component integrated into 6D object pose estimation networks to enhance their performance.

The qualitative and quantitative results of point cloud completion are presented in Fig. \ref{cover}(c) and Table \ref{point_cloud}, respectively. When incorporating surface normal information, the F1-score increases by 3\% compared to using point cloud information only. Additionally, the Chamfer distance \cite{butt1998optimum} based on the L1-norm (abbreviated as CD-$l_1$) and L2-norm (abbreviated as CD-$l_2$) decrease by 0.40 mm and 0.05 mm, respectively. These results indicate that our proposed 3F2N+ has a stronger ability to recover details in the point cloud completion task.

\begin{table}[t!]
    \settablefont
    \begin{center}
    \setlength{\tabcolsep}{6pt}
    \caption{Comparisons with respect to different input data for 6D object pose estimation.}
    \setlength\arrayrulewidth{0.5pt}
    \begin{tabular}{l|ccc}
        \hline
        Method
        &w/ 3F2N+
        &Accuracy ($\%$)
        &ADD (mm)\\
        \hline
        \multicolumn{1}{l|}{\multirow{2}*{PoseNet}}
        &
        &91.920
        &1.058
        \\       
        & \checkmark
        & \textbf{93.414}
        & \textbf{1.009}\\
        \hline
        \multicolumn{1}{l|}{\multirow{2}*{GCN}}
        &
        &98.284
        &6.067\\
        & \checkmark
        & \textbf{99.398}
        & \textbf{5.397}\\
        \hline
    \end{tabular}
    \label{6d_acc_dis}
    \end{center}
    \end{table}

\begin{table}[t!]
    \settablefont
    \begin{center}
    \setlength{\tabcolsep}{8pt}
    \caption{F1-score, CD-$l_1$, and CD-$l_2$ comparisons w/ and w/o 3F2N+ incorporated for point cloud completion.}
    \setlength\arrayrulewidth{0.8pt}
    \begin{tabular}{c|ccc}
        \hline
        w/ normal
        &F1-score (\%)
        &CD-$l_1$ (mm)
        &CD-$l_2$ (mm)\\

        \hline
        &0.497
        &11.622
        &0.577
        \\

        \checkmark
        & \textbf{0.511}
        & \textbf{11.221}
        & \textbf{0.530}\\
        \hline
    \end{tabular}
    \label{point_cloud}
    \end{center}
    \end{table}

\section{CONCLUSION}
\label{sec.conclusion}
    In this study, we introduced 3F2N+, an extension of 3F2N, enhanced with novel discontinuity discrimination techniques. These techniques leverage depth curvature minimization and correlation coefficient maximization to effectively address surface normal estimation challenges near or on discontinuities. To evaluate the robustness of SNEs on noisy data, we created a large-scale dataset, referred to as SSN, which contains both clean and noisy depth images. Through extensive experiments on the 3F2N and SSN datasets, we demonstrated the superior performance of 3F2N+ compared to existing SNEs. In addition, we explored the versatility of 3F2N+ in data-fusion freespace detection, 6D object pose estimation, and point cloud completion tasks. We believe DDM can serve as a universal solution for enhancing SNEs. Our commitment to ongoing research and development in this field remains unwavering.

\bibliographystyle{IEEEtran}


\end{document}